\documentclass[nohyperref]{article}

\usepackage{microtype}
\usepackage{graphicx}
\usepackage{subfigure}
\usepackage{booktabs} 

\usepackage{hyperref}

\usepackage[accepted]{icml2022}

\usepackage{amsmath}
\usepackage{amssymb}
\usepackage{mathtools}
\usepackage{amsthm}
\usepackage{multirow}
\usepackage{pifont}
\newcommand{\cmark}{\ding{51}}%
\newcommand{\xmark}{\ding{55}}%

\usepackage[capitalize,noabbrev]{cleveref}

\theoremstyle{plain}

\theoremstyle{definition}

\theoremstyle{remark}

\usepackage[disable,textsize=tiny]{todonotes}

\icmltitlerunning{Temporal Self-supervision for Video Transformers}

\begin{document}

\twocolumn[
\icmltitle{Time Is MattEr: Temporal Self-supervision for Video Transformers}




\icmlsetsymbol{equal}{*}

\begin{icmlauthorlist}
\icmlauthor{Sukmin Yun}{yyy}
\icmlauthor{Jaehyung Kim}{yyy}
\icmlauthor{Dongyoon Han}{comp}
\icmlauthor{Hwanjun Song}{comp}
\icmlauthor{Jung-Woo Ha}{comp}
\icmlauthor{Jinwoo Shin}{yyy,sch}
\end{icmlauthorlist}

\icmlaffiliation{yyy}{School of Electrical Engineering, KAIST, South Korea}
\icmlaffiliation{comp}{NAVER AI Lab, South Korea}
\icmlaffiliation{sch}{Graduate School of AI, KAIST, South Korea}

\icmlcorrespondingauthor{Sukmin Yun}{sukmin.yun@kaist.ac.kr}

\icmlkeywords{Machine Learning, ICML}

\vskip 0.3in
]



\printAffiliationsAndNotice{}  

\begin{abstract}

Understanding temporal dynamics of video is an essential aspect of learning better video representations. 
Recently, transformer-based architectural designs have been extensively explored for video tasks
due to their capability to capture long-term dependency of input sequences.
However, we found that
these Video Transformers are still biased to learn spatial dynamics rather than temporal ones,
and debiasing the spurious correlation is critical for their performance.
Based on the observations,
we design simple yet effective 
self-supervised tasks
for video models to learn temporal dynamics better.
Specifically, for debiasing the spatial bias,
our method learns the temporal order of video frames as extra self-supervision and enforces the randomly shuffled frames to have low-confidence outputs. 
Also,
our method learns the temporal flow direction of video tokens among consecutive frames for 
enhancing the correlation toward temporal dynamics.
Under various video action recognition tasks, we demonstrate the effectiveness of our method and its compatibility with state-of-the-art Video Transformers.
\end{abstract}

\section{Introduction}\label{intro}

Understanding videos for action or event recognition is a challenging yet crucial task that has gotten significant attention in computer vision communities \cite{lin2019tsm, cheng2021mask2former}.
Compared to image data, temporal dynamics between video frames provide additional information that is essential for recognition, as the actions and events generally occur over multiple consecutive frames. 
Hence, designing video-specific architectures for capturing such temporal dynamics has been a common theme in learning better video representations \cite{simonyan2014two, tran2015learning, tran2018R(2+1)D, feichtenhofer2019slowfast}. Recently, Transformer-based \cite{vaswani2017attention} video models, so-called \textit{Video Transformers}, have been extensively explored due to their capability to capture long-term dependency among the input sequence; for example, \citet{bertasius2021times} and \citet{patrick2021motions} introduce divided space-time and trajectory attentions, respectively.  

\begin{figure*}[t]
\begin{center}
    {
    \subfigure[Spurious correlation on spatial dynamics]
        {
        \includegraphics[width=0.31\textwidth]{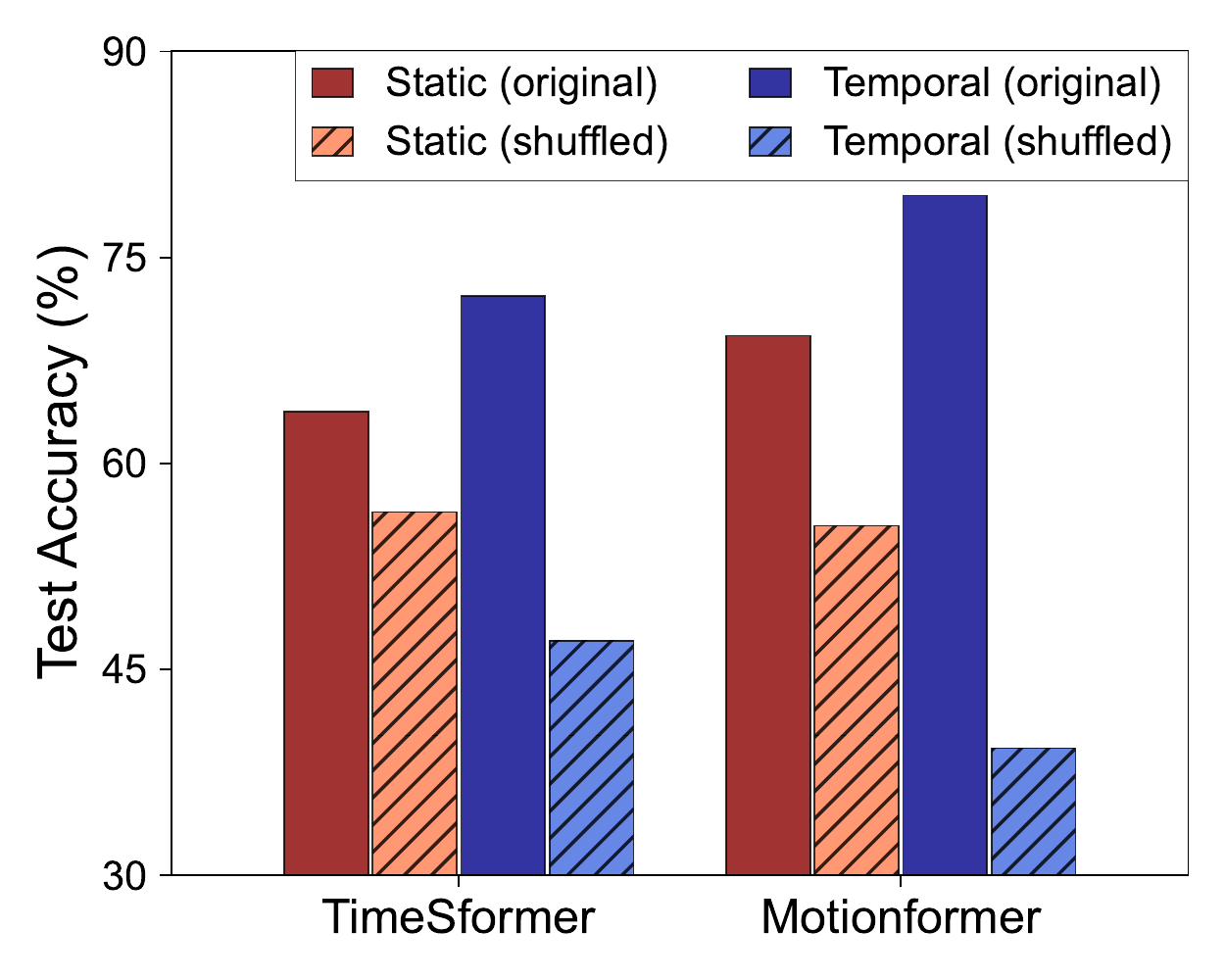}
        \label{fig:fig1a}
        }
    \subfigure[Vanishing temporal information]
        {
        \includegraphics[width=0.31\textwidth]{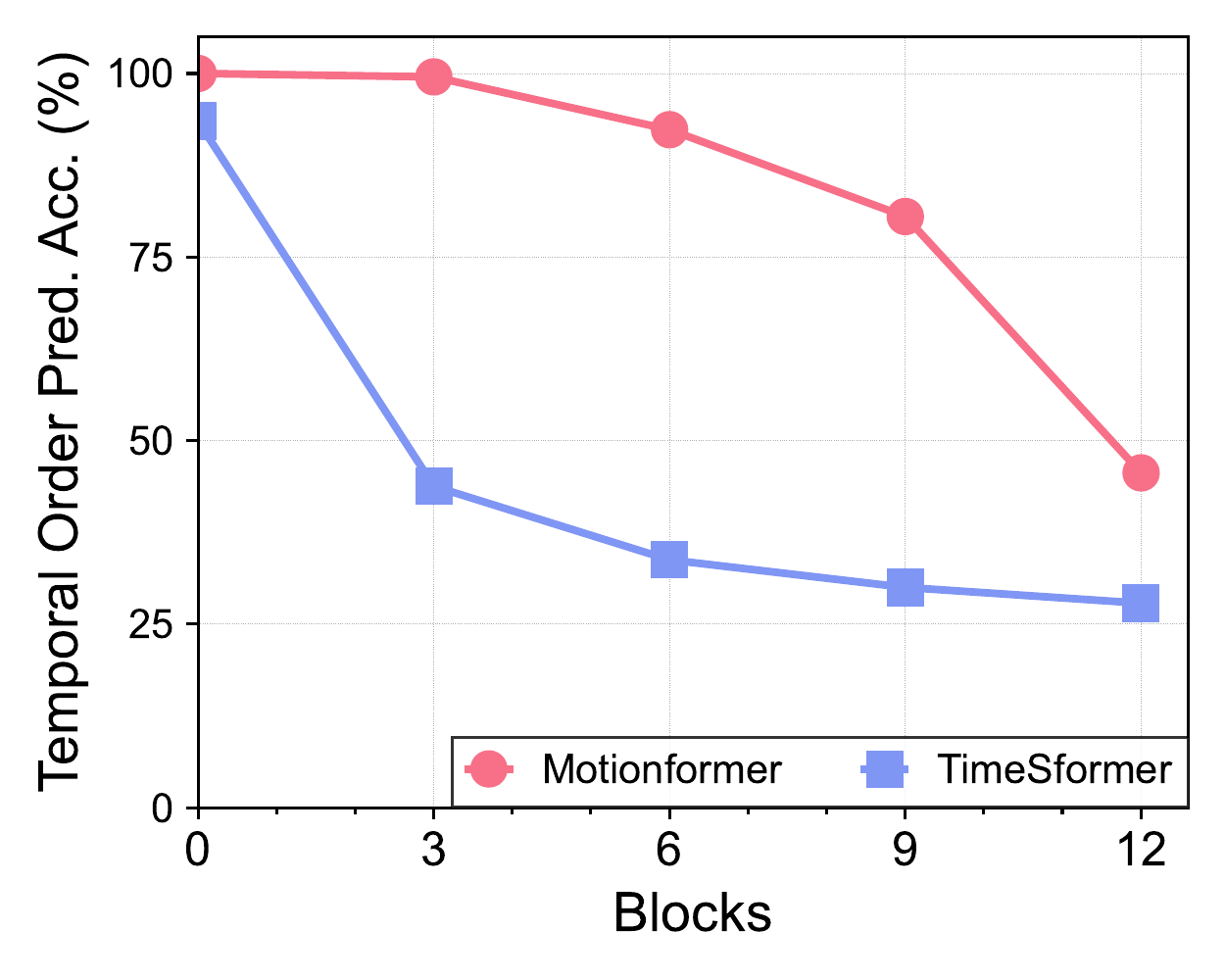}
        \label{fig:fig1b}
        }
    \subfigure[{Debiasing via temporal self-supervision}]
        {
        \includegraphics[width=0.31\textwidth]{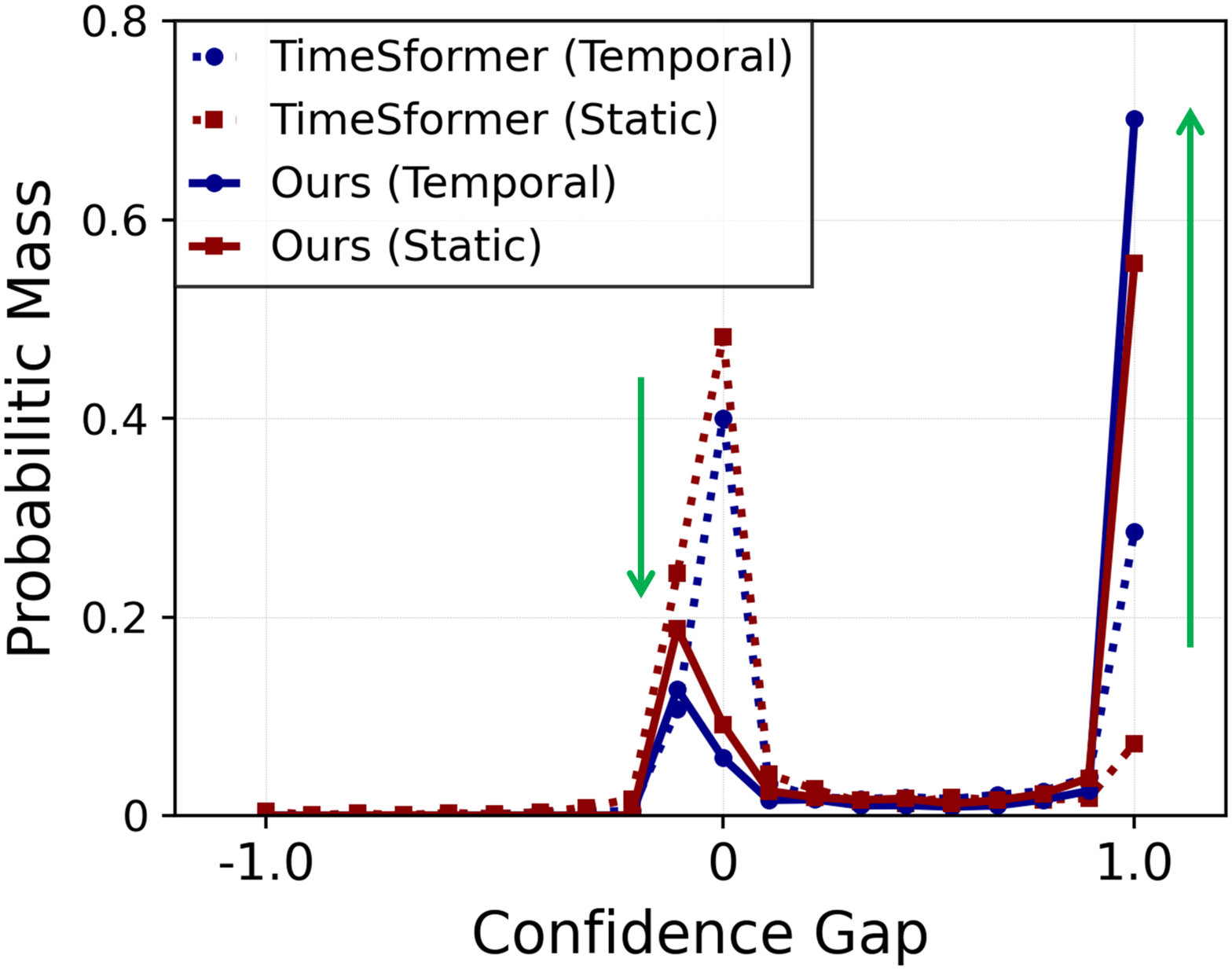}
        \label{fig:fig1c}
        }
    }
\end{center}
\caption{{Experimental results on SSv2 test dataset supporting our motivation. (a) Comparison of the accuracy for original and shuffled videos, respectively. Two different types of classes, \emph{Static} and \emph{Temporal}, are used to consider the relatively different importance of temporal information, following \citet{sevilla2021only}. 
Here, high accuracy is retained after shuffling frames, due to the models' bias toward spatial dynamics.
(b) Variation of temporal information within each Transformer block, measured with the accuracy of temporal order prediction; temporal information vanishes as the blocks deeper. (c) Probabilistic mass of the subset samples based on their confidence gap between original and shuffled videos. 20 bins are used to calculate the probabilistic mass. One could observe that the overall confidence gap is significantly increased from the proposed temporal self-supervised tasks, \emph{i.e.}, the model is successfully debiased.}}
\label{fig:fig1}
\end{figure*}

However, it is still questionable \emph{whether these architectural advances are enough to fully capture the temporal dynamics in a video.}
For example, \citet{fan2021image} shows that a well-designed image classifier with a flattened video along time dimension could outperform the several state-of-the-art video models on the representative action recognition tasks.
As additional clues, we observed that Video Transformers often predict a video action correctly with high confidence even when input video frames are randomly shuffled, \emph{i.e.,} the shuffled video does not contain correct temporal dynamics (see Figure \ref{fig:fig1a} and \ref{fig:fig1c}). Furthermore, as shown in Figure \ref{fig:fig1b}, Video Transformers also fail to capture the temporal order of video frames as their layers go deeper.
These observations reveal that the recent Video Transformers are likely to be biased to learn spatial dynamics, despite their efforts of designing a better architecture for learning the temporal one. 
Hence, this limitation inspires us to investigate an independent and complementary direction other than architectural advance, to improve the quality of learned video representations via better temporal modeling.

\begin{figure*}[t]
  \centering
  \includegraphics[width=0.99\textwidth]{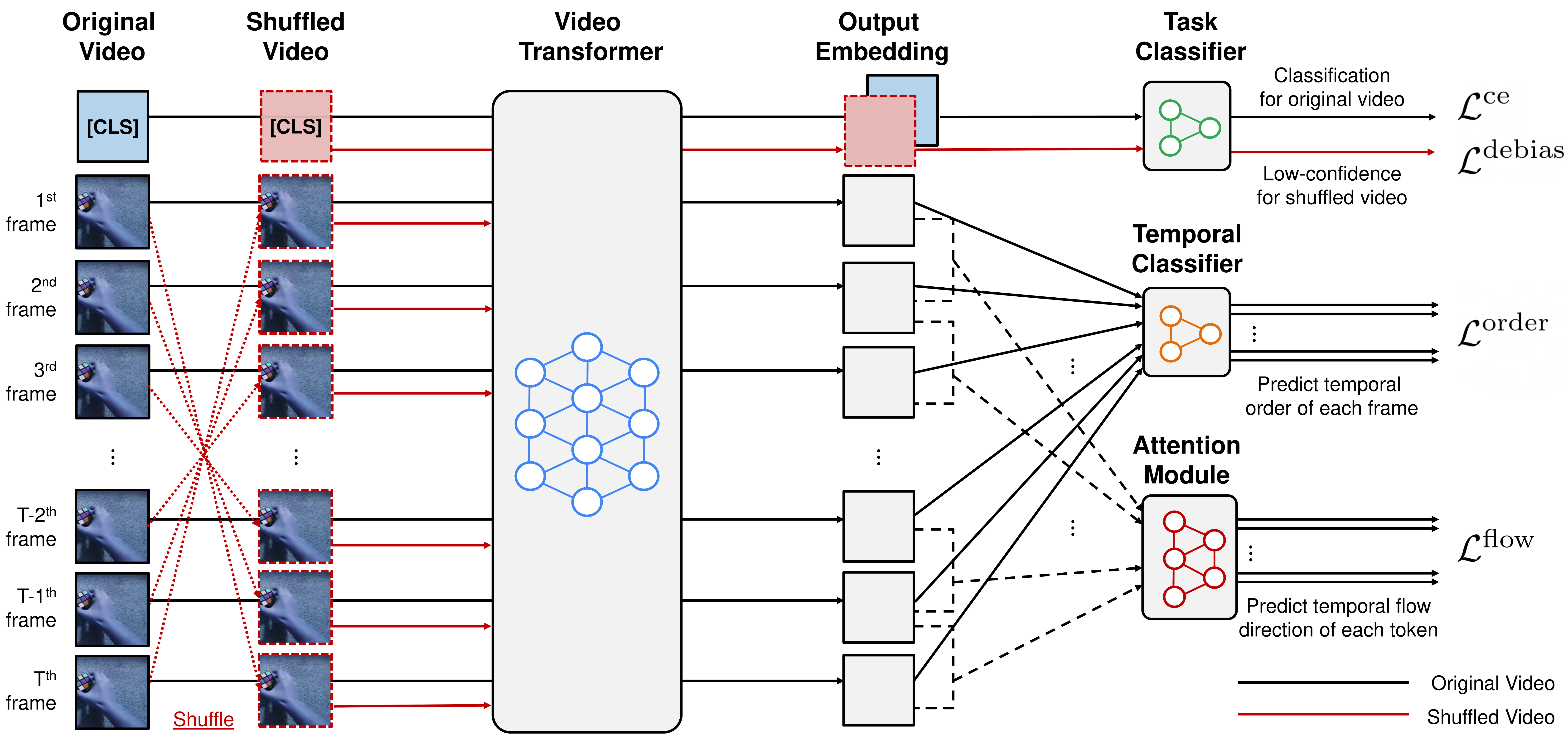}
  \caption{Illustration of the proposed scheme, TIME. 
  We use three types of temporal self-supervision for learning better video representations by (a) reducing a risk of learning the spurious correlation from spatial dynamics (\emph{i.e.,} $\mathcal{L}^{\text{debias}}$), (b) keeping temporal order information in deeper blocks (\emph{i.e.,} $\mathcal{L}^{\text{order}}$), and (c) enhancing the correlation toward temporal dynamics (\emph{i.e.,} $\mathcal{L}^{\text{flow}}$). Specifically, we train the model to (a) output low-confident predictions for shuffled videos, (b) predict the correct frame order of videos, and (c) predict nine types of temporal flow direction (eight angular directions and the center; see Eq.(\ref{eq:flow})) of video tokens in the consecutive frames.}\label{fig:fig2}
\end{figure*}

\noindent{\bf Contribution.} 
In this paper, we design simple yet effective 
\emph{frame-} and \emph{token-level}
self-supervised tasks, coined TIME (\textbf{T}ime \textbf{I}s \textbf{M}att\textbf{E}r), for video models which learn temporal dynamics better.
First, we train the models to learn two frame-level tasks for debiasing the spurious correlation learned from spatial dynamics.
Specifically, (a) we keep the temporal information within the video frame-by-frame by assigning the correct frame order as a self-supervised label to predict the temporal order of video frames (to avoid losing the temporal information), 
and (b) we simultaneously train the video models 
to be not able to output high-confident predictions when the input video does not contain the correct temporal order, \emph{i.e.,} randomly shuffled video frames.
Moreover, we train the models with a token-level task 
for enhancing the correlation toward temporal dynamics
by predicting the temporal flow direction of video tokens among consecutive frames.
To be specific, we adapt an attention-based module 
on the final representations of tokens in consecutive frames to predict their nine types of flow direction in the time axis (eight angular directions and the center; see Eq.(\ref{eq:flow})) by assigning their self-supervised labels obtained by 
Gunnar Farnebäck's algorithm~\cite{farneback2003two}.
We provide an overall illustration of our scheme in Figure~\ref{fig:fig2}.
It is worth noting that our scheme can be adopted to any Video Transformers in a plug-in manner and is beneficial to various video downstream tasks including action recognition
without additional human-annotated supervision. 
Somewhat interestingly,
our approach also
can be extended to the image domain for alleviating background bias.

To demonstrate the effectiveness of the proposed temporal self-supervised tasks, we incorporate our method with various Video Transformers and 
mainly evaluate on 
Something-Something-v2 (SSv2) \cite{goyal2017something} benchmark.\footnote{As \citet{sevilla2021only} stated, Something-Something-v2~\cite{goyal2017something} is known to contain a larger proportion of temporal classes requiring temporal information to be recognized. On the other hand, Kinetics~\cite{kay2017kinetics}
is not the case, where it is arguably much less suitable for modeling temporal dynamics well, \emph{e.g.,} see Figure \ref{fig:fig_kinetic} and Section \ref{experiments} for more details.
Hence, our method should have marginal gains on Kinetics.}
Despite its simplicity, our method consistently improves the recent state-of-the-art Video Transformers by debiasing the spurious correlation between spatial and temporal dynamics successfully. 
For example, ours improves the accuracy of TimeSformer \cite{bertasius2021times} and X-ViT \cite{bulat2021xvit} from 62.1\% to 63.7\% (+1.6\%) and 60.1\% to 63.5\% (+3.4\%) for SSv2, respectively.
Furthermore, we also found that our self-supervised idea of debiasing can be naturally extended to image classification models to reduce the spurious correlation learned from the image backgrounds, \emph{e.g.,}
our method improves the model generalization and robustness on ImageNet-9 
\cite{xiao2021noise}. For example, our idea not only improves DeiT-Ti/16~\cite{touvron2020training} from 77.3\% to 83.3\% (+6.0\%) on the original dataset but also from 50.3\% to 58.9\% (+8.6\%) on the Only-FG (\emph{i.e.,} remaining foregrounds only and removing backgrounds) in the background shift benchmarks \cite{xiao2021noise}.

Overall, our work highlights the importance of debiasing the spurious correlation of visual transformer models
with respect to the temporal or spatial dynamics. 
We believe our work could inspire researchers to rethink the under-explored, yet important problem 
and provide a new research direction.

\section{Related Work}\label{related}

\noindent{\bf Architectural advances for video action recognition.}
3D convolutional neural networks (3D-CNNs) were originally considered to learn deep video representations by inflating pre-trained ImageNet weights \cite{carreira2017I3D}. 3D-CNN models \cite{tran2018R(2+1)D, feichtenhofer2019slowfast} extract spatio-temporal representations via their own temporal modeling methods; for example, SlowFast~\cite{feichtenhofer2019slowfast} captures short- and long-range of time dependencies by using two different speed of pathways for the video. However, such 3D convolutional designs are limited to capture long-term dependency of video with its small receptive field.

Due to the ability to capture long-term dependency of the self-attention mechanism, transformer-based models \cite{neimark2021VTN, bertasius2021times, arnab2021vivit, patrick2021motions, bulat2021xvit,fan2021image, li2022uniformer} have been recently explored for video action recognition 
by following the success of Vision Transformer \cite{dosovitskiy2021vit}, which has shown competitive performances against CNNs in image domains. 
In particular, Video Transformers such as TimeSformer \cite{bertasius2021times} and ViViT \cite{arnab2021vivit} propose the use of a temporal attention module with the existing Vision Transformer to better understand temporal dynamics.

Besides, several works attempt to develop a more efficient and powerful attention module to mitigate the quadratic complexity of the self-attention in learning from videos.
For example, TimeSformer suggests divided-space-time attention module by decomposing the time and space dimensions separately, 
X-ViT~\cite{bulat2021xvit} further restricts time attention to a local temporal window, \emph{i.e.,} local space-time attention. Motionformer~\cite{patrick2021motions} introduces a trajectory attention to model the probabilistic
path of a token between frames over the entire space-time feature volume.

\textbf{Static biases in video.}
In video action recognition, it is essential capturing long-term dependency of temporal dynamics. 
However, several prior works \cite{li2019repair,sevilla2021only,huang2018whatmakesvideo} have shown that video models are often biased to learn spatial dynamics rather than the temporal one due to the presence of static classes, which contain informative frames to predict the action class labels without understanding overall temporal information. 
In particular, \citet{sevilla2021only} 
categorizes action classes in video datasets as temporal and static classes with respect to whether they necessarily require understanding temporal information to recognize them, and shows that training on temporal classes can lead video models to avoid spatial bias and generalize better.

In the end, most recent works have mainly focused on architectural advances to capture temporal dynamics in videos. On the contrary, our method aims to handle this issue by designing self-supervised tasks 
not only for capturing stronger temporal dynamics, but also for debiasing the spurious correlation learned from spatial dynamics.
Meanwhile, some CNN-based works~\cite{misra2016shuffle, lee2017unsupervised, hu2021contrast} have also adapted a binary classification task that predicts the correct order of randomly chosen video frames or clips, 
however, our idea is simple and computationally efficient for Video Transformers; their capabilities to capture long-term dependency enable us to infer all the absolute temporal order frame-by-frame, while the prior works under CNNs do the binary order one-by-one.

\section{Motivation: Bias toward Spatial Dynamics}\label{sec:example}

\begin{table*}[t]
\centering
\scalebox{0.95}{
\begin{tabular}{lccccccc}
\toprule 
Model & Input frame & Tokenization & Top-1 {\color{blue}$\uparrow$} & Top-5 {\color{blue}$\uparrow$} & Shuffled Top-1 {\color{red}$\downarrow$} & Shuffled Top-5 {\color{red}$\downarrow$}  \\
\midrule
TimeSformer & 8 & $1\times16\times16$ & 59.1 & 85.6 & 46.4 & 78.8\\
TimeSformer-HR & 16 & $1\times16\times16$ & 61.8 & 86.9 & 49.7 & 81.5\\
TimeSformer-L & 64 & $1\times16\times16$ & 62.0 & 87.5 & 55.1 & 84.7\\
\midrule
Motionformer & 16 & $2\times16\times16$ & 66.5 & 90.1 & 43.9 & 75.6\\
Motionformer-L & 32 & $2\times16\times16$ & 68.1 & 91.2 & 40.7 & 73.3\\
\bottomrule
\end{tabular}}
\caption{Evaluation of pre-trained Video Transformers on SSv2 dataset. Top-1 and Top-5 denote top-1 and top-5 test accuracy with the original input video, respectively. Shuffled Top-1 and Shuffled Top-5 denote the accuracy of frame-shuffled videos, respectively.}
\label{table1}
\end{table*}

\subsection{Preliminaries: Video Transformers}\label{sec:method:pre}

Let $\mathbf{x}=[\mathbf{x}_1, \dots, \mathbf{x}_T]\in\mathbb{R}^{T\times H\times W \times C}$ be an input video 
where 
$(H, W)$ is the spatial resolution,
$T$ is the number of frames, 
and $C$ is the number of channels.
Video Transformers (\emph{e.g.,} TimeSformer \cite{bertasius2021times}) process the input video $\mathbf{x}$ as a sequence of $hwt$ video tokens $\{\mathbf{x}^{(i)}\in\mathbb{R}^{(T/t) \times (H/h) \times (W/w) \times C}\}_{i=1}^{hwt}$
and then linearly transform them to $D$-dimensional embeddings $\mathbf{e}^{(i)}=E\mathbf{x}^{(i)}+E_{\text{pos}}^{(i)}\in\mathbb{R}^D$,
where $E \in \mathbb{R}^{D \times (THWC / thw)}$
is the linear transformation and $E_{\text{pos}}^{(i)}\in\mathbb{R}^{D}$ is the positional embedding for the $i$-th token $\mathbf{x}^{(i)}$. 
Following \citet{dosovitskiy2021vit}, 
Video Transformers also utilize the $\mathtt{[CLS]}$ token $\mathbf{e}^\mathtt{[CLS]}\in\mathbb{R}^D$ to represent 
the entire sequence of video tokens (\emph{i.e.,} the input video $\mathbf{x}$) 
by prepending it to the token embedding sequence as
$\mathbf{e}=[\mathbf{e}^\mathtt{[CLS]};\mathbf{e}^{(1,1)};\ldots;\mathbf{e}^{(s,t)}]$, 
where $s=hw$.
Then, Video Transformers take the sequence $\mathbf{e}$ as inputs and then output the same size contextualized embeddings with their own spatial-temporal attention module.

\noindent{\bf Spatio-temporal attentions.} 
Spatio-temporal attention is an extension of the self-attention that operates over space and time dimensions in parallel. 
For a video input sequence $\mathbf{e}\in\mathbb{R}^{st \times D}$ with a space-time location $st$,
joint spatio-temporal attention is a natural extension that can be defined:
\begin{align}
    \mathbf{q}_{st} = \mathbf{e}&Q, \mathbf{k}_{st} = \mathbf{e}K, \mathbf{v}_{st} = \mathbf{e}V, \\
    \text{(joint space-time)} & \sum_{s^\prime t^\prime}\mathbf{v}_{s^\prime t^\prime}\cdot
    \frac{\exp{\langle\mathbf{q}_{st}, \mathbf{k}_{s^\prime t^\prime}\rangle}}{\sum_{\bar{s}\bar{t}}\exp{\langle\mathbf{q}_{st}, \mathbf{k}_{\bar{s}\bar{t}}\rangle}},
\end{align}
where $Q, K, V \in\mathbb{R}^{D \times D}$ are query, key, and value matrices.
Here, the joint attention computes attentions of all keys $\mathbf{k}_{s^\prime t^\prime}$ for each query $\mathbf{q}_{st}$, 
and it has a limitation of quadratic complexity in both space and time, \emph{i.e.,} $\mathcal{O}(s^2 t^2)$.
To address this limitation, several Video Transformers \cite{bertasius2021times,arnab2021vivit} propose \textit{divided attention}, which restricts space or time dimension as below:
\begin{align}
    \text{(space only)} \:\:\: & \sum_{s^\prime}\mathbf{v}_{s^\prime t}\cdot
    \frac{\exp{\langle\mathbf{q}_{st}, \mathbf{k}_{s^\prime t}\rangle}}{\sum_{\bar{s}}\exp{\langle\mathbf{q}_{st}, \mathbf{k}_{\bar{s}t}\rangle}}, \\
    \text{(time only)} \:\:\: & \sum_{t^\prime}\mathbf{v}_{s t^\prime}\cdot
    \frac{\exp{\langle\mathbf{q}_{st}, \mathbf{k}_{s t^\prime}\rangle}}{\sum_{\bar{t}}\exp{\langle\mathbf{q}_{st}, \mathbf{k}_{s\bar{t}}\rangle}}.
\end{align}
The divided attention reduces the complexity to $\mathcal{O}(s^2 t) + \mathcal{O}(s t^2)$, and TimeSformer~\cite{bertasius2021times} and ViViT~\cite{arnab2021vivit} utilize this approach alternately for getting spatio-temporal features.
On the other hand, Motionformer~\cite{patrick2021motions} also introduces trajectory attention, which is designed to capture temporal dynamics by modeling a set of trajectory tokens computed across the frames, with a complexity of $\mathcal{O}(s^2 t^2)$.

In order to be concise,
we use $f_\theta$ to denote the encoder of Video Transformers parameterized by $\theta$
as follows:\footnote{Note that $\theta$ contains all Video Transformer parameters, including the encoder $f_\theta$ and the linear classifier head $g_\theta$.}
\begin{align}
f_\theta(\mathbf{x})
&=f_\theta([\mathbf{e}^\mathtt{[CLS]};\mathbf{e}^{(1,1)};\ldots;\mathbf{e}^{(s,t)}])\nonumber\\
&:=[f_\theta^\mathtt{[CLS]}(\mathbf{x});f_\theta^{(1,1)}(\mathbf{x});\ldots;f_\theta^{(s,t)}(\mathbf{x})],\label{eq:vit}
\end{align}
where $f_\theta^\mathtt{[CLS]}(\mathbf{x})$ and $f_\theta^{(i,j)}(\mathbf{x})$ are the final representations of the $\mathtt{[CLS]}$ token and the $(i,j)$-th token, respectively. 
We remark that $f_\theta^\mathtt{[CLS]}(\mathbf{x})$ is generally utilized for solving video-level downstream tasks such as action recognition with a linear classifier head $g_\theta$.

\subsection{Observations}\label{observe}

In this section, we describe our empirical observations based on the recent Video Transformers, such as TimeSformer \cite{bertasius2021times} and Motionformer \cite{patrick2021motions}, trained to recognize the actions in video using SSv2 dataset \cite{goyal2017something}. 
Here, our observations reveal that even the recent state-of-the-art video models still struggle to fully exploit the temporal information in video data. 
These findings serve as a key intuition for designing our temporal self-supervised tasks for video models.

\noindent{\bf Spurious correlation on spatial dynamics.} 
Our first observation is that the violation of temporal dynamics within video does not lead to the low confident predictions of Video Transformers. 
Intuitively, if the models are learned to recognize the actions via capturing the temporal dynamics between the frames, their predictions should have low confidence when the input video does not have the correct temporal order, \emph{e.g.}, frames are randomly shuffled \cite{misra2016shuffle, sevilla2021only}. 
To verify such behavior, we measure the accuracy of Video Transformers on SSv2 test dataset with original and shuffled videos in Table \ref{table1};
the shuffled video $\widetilde{\mathbf{x}}$ is constructed from the original video $\mathbf{x}=[\mathbf{x}_1, \dots, \mathbf{x}_T]$ where $\widetilde{\mathbf{x}}=[\mathbf{x}_{p(1)}, \dots, \mathbf{x}_{p(T)}]$ and $p(i)$ is a permuted order.
Here, it is observed that the models generally achieve the high test accuracy regardless of the shuffling of videos; 
for example, under 64 frames video, TimeSformer achieves 62.0\% accuracy with only 6.9\% reduction compared to the accuracy of the original video.
We note that such high confident predictions on the violation of temporal dynamics (\emph{i.e.,} incorrect temporal order) often occur even for \emph{temporal classes} where temporal information is crucial to recognize them \cite{sevilla2021only} (see Figure \ref{fig:fig1a}).
These results indicate that the Video Transformers are likely to be biased to learn the spatial dynamics rather than temporal one despite their specific architectural designs to learn temporal information better.\footnote{Such spatial bias in video models may come from the ImageNet pre-trained weights. 
However, even without the pre-trained weights, we empirically found that our method still achieves significant improvements in training from scratch (see Appendix \ref{app:scratch}).}

\begin{table}[t]
\centering
\scalebox{0.95}{
\begin{tabular}{lcc}
\toprule
Model  & Top-1 & Top-5 \\ 
\midrule
TimeSformer \cite{bertasius2021times} & 62.1 & 86.4 \\
TimeSformer + TIME  & \textbf{63.7} & \textbf{87.8}\\
\cmidrule(lr){1-3}
Motionformer \cite{patrick2021motions} & 63.8 & 88.5\\
Motionformer + TIME & \textbf{64.7} & \textbf{89.3}\\
\cmidrule(lr){1-3}
X-ViT \cite{bulat2021xvit} & 60.1 & 85.2\\
X-ViT + TIME  & \textbf{63.5} & \textbf{88.1} \\
\bottomrule
\end{tabular}
}
\caption{Video action recognition performance of the recent Video Transformers. All models share the same training details, and they are fine-tuned on the SSv2 dataset from the ImageNet-1k pre-trained weights. Top-1 and -5 denote test accuracies, respectively.}\label{table:ssv2}
\end{table}

\noindent{\bf Vanishing temporal information.} 
Next, we further observe that 
the deeper the transformer blocks in Video Transformers even fail to keep the temporal order of video frames.
Specifically, we first generate temporal labels $\mathbf{y}_{\text{order}}^{(j)}=j$ as the temporal order of $j$-th frames. 
Then, we train additional linear classifier ${g}^{\text{order}}_{\theta}$ using $\mathcal{L}^{\text{order}}$ to predict $\mathbf{y}_{\text{order}}^{(j)}$ based on the frozen output embeddings of input video $\mathbf{x}$ as follow:
\begin{align}
\bar{f}_\theta^{(j)}(\mathbf{x}) &:= \frac{1}{s}\sum_{i=1}^{s}f_\theta^{(i,j)}(\mathbf{x}),\\
\mathcal{L}^{\text{order}}(\mathbf{x})&:= \frac{1}{t}\sum_{j=1}^t \text{CE}({g}^{\text{order}}_\theta(\bar{f}_\theta^{(j)}(\mathbf{x})),\mathbf{y}_{\text{order}}^{(j)}),\label{eq_tempo}
\end{align}
where $\bar{f}_\theta^{(j)}(\mathbf{x})$ is an aggregated embedding of $f_\theta^{(i,j)}(\mathbf{x})$ across the space axis, and $\text{CE}(\mathbf{x}, \mathbf{y})$ denotes the standard cross-entropy loss between an input $\mathbf{x}$ and a given label $\mathbf{y}$, respectively. 
To track the change of temporal information within each transformer block, 
we train a linear classifier $g^{\text{order}}_\theta$ on the aggregated embedding 
of each block
and compare the accuracy of the temporal order prediction. 
As shown in Figure \ref{fig:fig1b}, the earlier blocks show much higher accuracy, but the performance significantly decreases in the latter blocks;
for example, Motionformer achieves 99.5\% accuracy at the 3th block, but it decreases to 45.6\% at the last block.
It might be the results of the learned spurious correlation of video models;
as the models are focused on learning the spatial information, the temporal information would be less captured and lost.

Overall, these empirical observations reveal that only designing better video model architecture may not be enough;
\footnote{Architectural modifications could be an alternative for keeping temporal information. 
Nevertheless, we empirically found that our method could further improve the performance (see Appendix \ref{app:arch_modify}).}
hence, it motivates us to investigate the independent yet complementary direction for improving the quality of learned video representation.

\begin{figure}[t]
  \centering
  \vspace{-1mm} 
  \includegraphics[width=0.7\columnwidth]{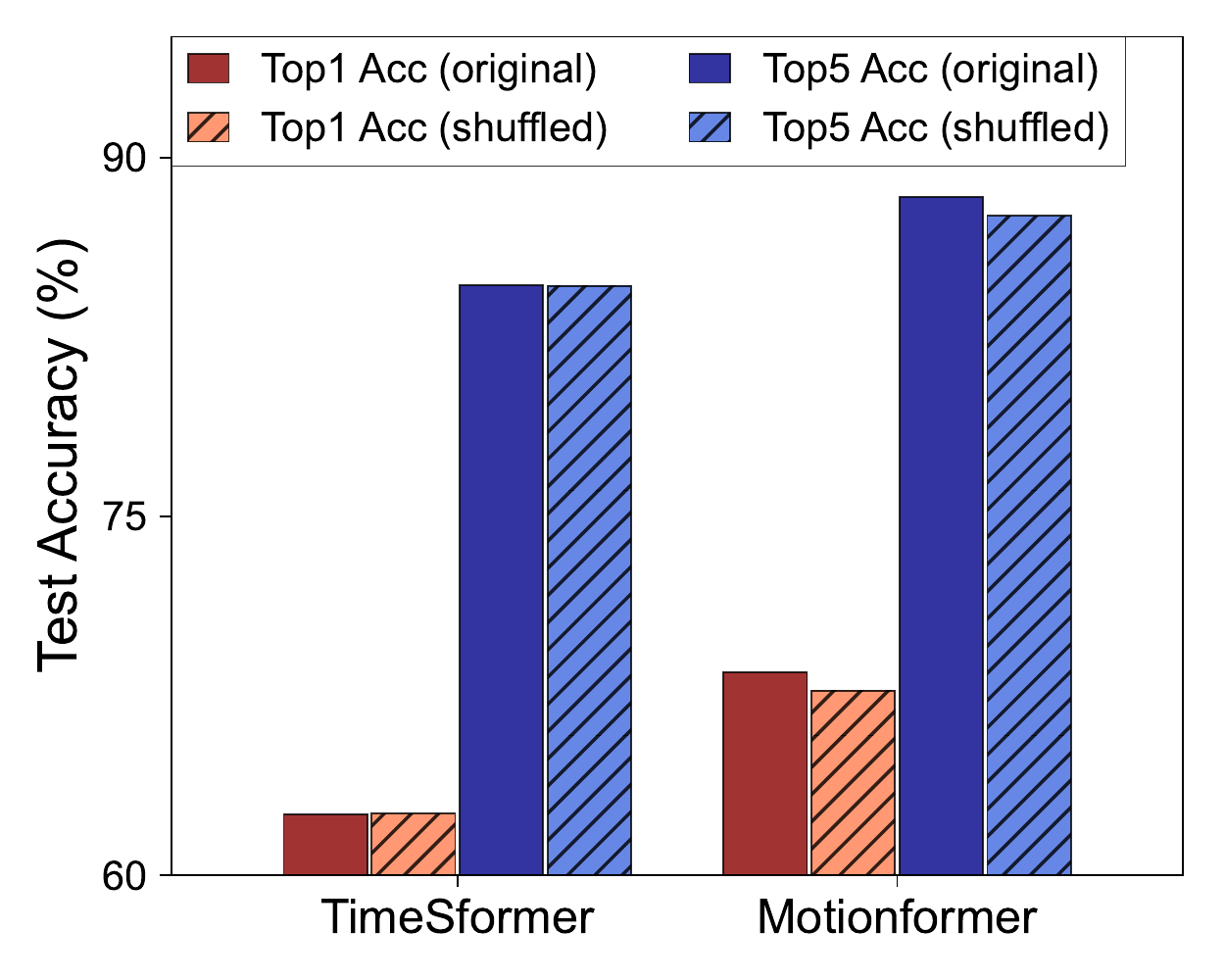}
  \vspace{-4mm} 
  \caption{Top-1 and top-5 test accuracy on Kinetics-400~\cite{kay2017kinetics} with original and shuffled videos, respectively.}\label{fig:fig_kinetic}
\end{figure}

\begin{figure*}[t]
  \centering
  \includegraphics[width=0.85 \textwidth]{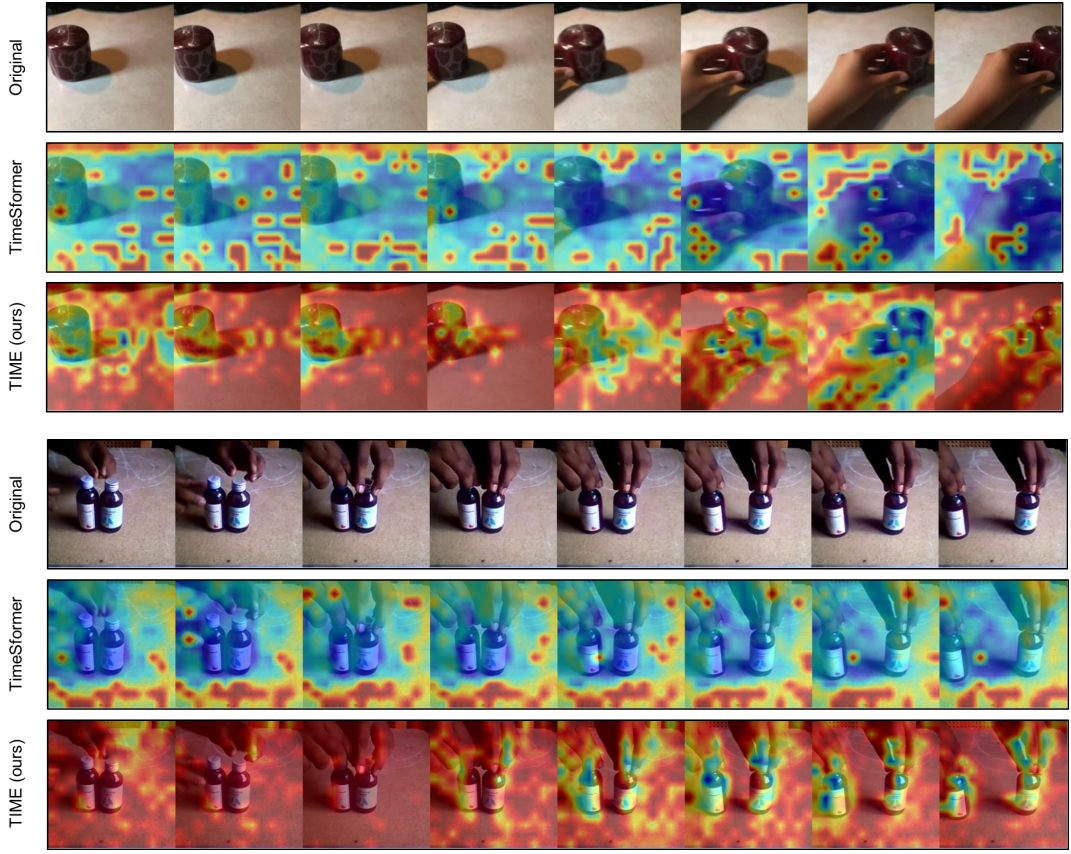}
  \caption{Visualization of learned video models via GradCAM \cite{selvaraju2017grad}. 
  Here, we present eight frame input videos that come from \emph{pushing something from left to right} class (\textbf{Top})
  and \emph{moving something and something away from each other} class (\textbf{Bottom}) in the SSv2 test dataset, respectively.
  Video models are fine-tuned on the SSv2 dataset from the ImageNet-1k pre-trained weights.
  While TimeSformer fails to focus on the object, TimeSformer + TIME (ours) is successfully tracking its trajectory. Best viewed in color.}\label{fig:fig3}
\end{figure*}

\begin{table*}[t]
\centering
\scalebox{0.95}{
\resizebox{\textwidth}{!}{
\begin{tabular}{cccccccccccc}
\toprule
& & & \multicolumn{3}{c}{{SSv2 dataset}} & \multicolumn{3}{c}{{Temporal subset}} & \multicolumn{3}{c}{{Static subset}} \\
\cmidrule(lr){4-6}\cmidrule(lr){7-9}\cmidrule(lr){10-12}
$\mathcal{L}^{\text{order}}$ & $\mathcal{L}^{\text{debias}}$ & $\mathcal{L}^{\text{flow}}$ & Original {\color{blue}$\uparrow$} & Shuffled {\color{red}$\downarrow$} & Gap {\color{blue}$\uparrow$} & Original {\color{blue}$\uparrow$} & Shuffled {\color{red}$\downarrow$} & Gap {\color{blue}$\uparrow$} & Original {\color{blue}$\uparrow$} & Shuffled {\color{red}$\downarrow$} & Gap {\color{blue}$\uparrow$}\\
\midrule
\xmark & \xmark & \xmark & 62.1 & 41.3 & 20.8 & 84.9 & 57.0 & 27.9 & 84.1 & 84.1 & 0.0\\
\cmark & \xmark & \xmark & 62.6 & 39.7 & 22.9 & 88.0 & 51.4 & 36.6 & 85.3 & \underline{83.2} & \underline{2.1}\\
\xmark & \cmark & \xmark & 62.7 & \textbf{10.9} & \textbf{51.8} & 88.1 & \underline{18.8} & \underline{69.5} & 84.6 & 84.5 & 0.1\\
\cmark & \cmark & \xmark & \underline{63.2} & 30.4 & 32.8 & \underline{90.0} & \textbf{18.6} & \textbf{71.4} & \underline{86.3} & \textbf{40.0} & \textbf{46.3}\\
\xmark & \xmark & \cmark & 62.6 & 39.5 & 23.1 & 87.3 & 56.4 & 30.9 & 84.4 & 84.3 & 0.1\\
\cmark & \xmark & \cmark & 62.7 & 40.7 & 22.0 & 88.6 & 52.5 & 36.1 & \underline{85.7} & 83.8 & 1.9\\
\xmark & \cmark & \cmark & \underline{63.4} & \underline{13.0} & \underline{50.4} & \underline{89.3} & 22.2 & 67.1 & 85.0 & 84.9 & 0.1\\
\cmark & \cmark & \cmark & \textbf{63.7} & \underline{25.3} & \underline{38.4} & \textbf{90.2} & \underline{22.1} & \underline{68.1} & \textbf{86.9} & \underline{69.3} & \underline{17.6}\\
\bottomrule
\end{tabular}}
}
\caption{Ablation study on effect of loss components $\mathcal{L}^{\text{order}}$, $\mathcal{L}^{\text{debias}}$ and $\mathcal{L}^{\text{flow}}$. All models share the same training details and are fine-tuned 
from ImageNet-1k pre-trained weights. 
``Original'' and ``Shuffled'' denote the top-1 accuracy of the original and shuffled input videos, respectively, and ``Gap'' denotes the difference between the Original and Shuffled accuracies. 
``SSv2 dataset'', ``Temporal subset,'' and ``Static subset'' denote 
training datasets that show relatively different importance of temporal information; 
configurations of the Temporal and Static subsets are reported in Appendix~\ref{app:temporal_static}.
The best scores are in \textbf{bold}, and the top 3 scores are \underline{underlined}.
}\label{table:lossweight}
\end{table*}

\section{TIME: Temporal Self-supervision for Video}\label{method}
Motivated by the previous observations in Section \ref{observe}, we introduce a simple yet effective self-supervised tasks, coined TIME (\textbf{T}ime \textbf{I}s \textbf{M}att\textbf{E}r), to better understand temporal dynamics of videos, which can be beneficial for video recognition in a model-agnostic way.
Overall illustration of the proposed scheme is presented in Figure \ref{fig:fig2}.

\noindent{\bf Debiasing spatial dynamics.} 
First, we reduce a risk of learning the spurious correlation
from spatial dynamics by utilizing the shuffled (\emph{i.e.,} temporarily incorrect) for training; we train the video models to output the low confident predictions
for the shuffled video.
Specifically, we minimize the Kullback-Leibler (KL) divergence from the predictive distribution on randomly shuffled video $\widetilde{\mathbf{x}}$ to the uniform one in order to give less confident predictions as follows:
\begin{align}
\mathcal{L}^{\text{debias}}(\widetilde{\mathbf{x}}):= 
\text{KL}\left(\mathcal{U}(\mathbf{y})
    \big\| \mathtt{Softmax}(g_\theta(f_\theta^\mathtt{[CLS]}(\mathbf{\widetilde{x}})))\right),
\end{align}
where KL is a the Kullback-Leibler (KL) divergence, $\mathtt{Softmax}$ is a softmax function, $g_\theta$ is a linear classification head, and $\mathcal{U}(\mathbf{y})$ is the uniform distribution.
As it prevents the model from biased predictions toward spatial dynamics,
the model learns to exploit the temporal dynamics better.
We remark that the similar approaches have been utilized in other domains, such as image \cite{lee2018training, hendrycks2019deep} and language \cite{moon2021masker}. 

\noindent{\bf Learning temporal order of frames.} 
To avoid losing temporal order information in deeper blocks, we directly regularize the model to keep such information at the final block.
Specifically, we simply add a linear classifier ${g}^{\text{order}}_\theta$ on the final aggregated embedding $\bar{f}_\theta^{(j)}(\mathbf{x})$ of each frame and train the model using $\mathcal{L}^{\text{order}}$ (\ref{eq_tempo}) to predict the temporal order of input video frames as described in Section \ref{observe}.
This regularization preserves the temporal information until the end of blocks; hence, the model could utilize it for solving the target task, such as action recognition.

\noindent{\bf Learning temporal flow direction of tokens.} 
To enhance the correlation toward temporal dynamics,
we train the model to predict the temporal flow direction of tokens in the consecutive frames
$\mathbf{x}_j$ and $\mathbf{x}_{j+1}$.
To be specific, we adopt an attention-based module $h_\theta$ on pairs of final representations of the consecutive frames
$\hat{f}^{(j;j+1)}(\mathbf{x})$,
and train the model using $\mathcal{L}^{\text{flow}}$ to predict their nine types of flow direction in the time axis
(eight angular directions and the center)
by assigning self-supervised labels 
$\mathbf{y}_{\text{flow}}^{(i,j)}$ obtained by 
Gunnar Farnebäck's algorithm~\cite{farneback2003two}
as follow:
\begin{align}
&\{r^{(i,j)}, \phi^{(i,j)}\}_i := \mathtt{Polar}\left(\mathtt{Farneb\ddot{a}ck}(\mathbf{x}_j, \mathbf{x}_{j+1})\right),
\\
&\mathbf{y}_{\text{flow}}^{(i,j)} := 
\begin{cases}
    \lfloor\phi^{(i,j)}\cdot\frac{4}{\pi}\rfloor + 1 & \text{if} \: r^{(i,j)}\geq \tau,\\
    0              & \text{otherwise},
\end{cases}\label{eq:flow}\\
&\hat{f}^{(j;j+1)}(\mathbf{x}):=(\{f_\theta^{(i,j)}(\mathbf{x})\}_i, \{f_\theta^{(i,j+1)}(\mathbf{x})\}_i), \\
&h_\theta(\hat{f}^{(j;j+1)}(\mathbf{x}))
:=[h_\theta^{(1,j)}(\mathbf{x});\ldots;h_\theta^{(s,j)}(\mathbf{x})],\\
&\mathcal{L}^{\text{flow}}(\mathbf{x}):= \frac{1}{s(t-1)}\sum_{j=1}^{t-1}\sum_{i=1}^s\text{CE}(h_\theta^{(i,j)}(\mathbf{x})),\mathbf{y}_{\text{flow}}^{(i, j)}),\label{eq:token}
\end{align}
where $r^{(i,j)}>0$ and $0\leq\phi^{(i,j)}<2\pi$ are the magnitude and angle obtained from the frames $\mathbf{x}_j$ and $\mathbf{x}_{j+1}$ via the polarization function $\mathtt{Polar}$ and the Gunnar Farnebäck's algorithm $\mathtt{Farneb\ddot{a}ck}$, and $\tau$ is a threshold for the center and angular directions of self-supervised labels $\mathbf{y}_{\text{flow}}^{(i,j)}$.

In summary, our total training loss can be written as follows:
\begin{align}
\mathcal{L}^{\text{TIME}}(\mathbf{x},\mathbf{\tilde{x}}):=&\lambda^{\text{order}}\mathcal{L}^{\text{order}}(\mathbf{x})+\lambda^{\text{debias}}\mathcal{L}^{\text{debias}}(\tilde{\mathbf{x}})\nonumber\\
&+\lambda^{\text{flow}}\mathcal{L}^{\text{flow}}({\mathbf{x}}),\label{eq:time}
\\
\mathcal{L}^{\text{total}}(\mathbf{x},\mathbf{\tilde{x}}):=&\mathcal{L}^{\text{ce}}(\mathbf{x})+\mathcal{L}^{\text{TIME}}(\mathbf{x}, \tilde{\mathbf{x}}),
\end{align}
where $\mathcal{L}^{\text{ce}}$ is the cross-entropy loss with a linear classification head $g_\theta$ for video recognition, $\lambda^{\text{order}}$, $\lambda^{\text{debias}}$ and $\lambda^{\text{flow}}$ are hyperparameters. We simply set all of them to be 1 in our experiments (see Section~\ref{sec:ablation} for analysis on $\lambda$).

\section{Experiments}\label{experiments}
In this section, we demonstrate the effectiveness of the proposed temporal self-supervision, TIME. 
Specifically, we incorporate TIME with the state-of-the-art Video Transformers~\cite{bertasius2021times,patrick2021motions,bulat2021xvit} 
and evaluate their temporal modeling ability on 
Something-Something-v2 (SSv2)~\cite{goyal2017something}, which contains a large proportion of temporal classes than other video dataset like Kinetics-400~\cite{kay2017kinetics}. 
We also use
temporal and static classes in the SSv2 dataset, stated by \citet{sevilla2021only}, for validating the importance of temporal information; 
temporal classes necessarily require temporal information for action recognition.
More details of experimental setups are described in each section.

\vspace{0.03in}
\noindent{\bf Video Datasets.} 
We use SSv2 \cite{goyal2017something} datasets and its temporal and static classes following the categorization of \citet{sevilla2021only} to evaluate whether video models understand the temporal dynamics well.
Notably, SSv2 is a challenging dataset that consists of $\sim$169k training videos and $\sim$25k validation videos over 174 classes; in particular, it 
contains a large proportion of temporal classes requiring temporal information to be recognized~\cite{sevilla2021only}.
To investigate behaviors of video models on relatively different importance of temporal information,
we further construct ``Temporal subset'' and ``Static subset'' as the benchmarks by choosing 6 temporal classes and 16 static classes from the above temporal and static classes, respectively.
We report the specific labels of the Temporal and Static subsets in Appendix \ref{app:temporal_static}, respectively.

Meanwhile, Kinetics dataset \cite{kay2017kinetics} (\emph{e.g.,} Kinetics-400) is a large-scale video dataset, which consists of $\sim$240k training videos and $\sim$20k validation videos in 400 human action categories. However, it arguably contains fewer temporal classes and 
is comprised of a large amount of static classes~\cite{li2019repair,huang2018whatmakesvideo,sevilla2021only,fan2021image}; 
For example, \citet{fan2021image} reports that changing temporal order of kinetic videos does not drop the recognition performance, and we also found similar observations on a 10\% subset of Kinetics 400; Figure \ref{fig:fig_kinetic} shows that
the state-of-the-art Video Transformers~\cite{bertasius2021times, patrick2021motions} have almost the same accuracy even their input video frames are randomly shuffled, unlike SSv2 in Figure~\ref{fig:fig1a}. 
To this end, we solely use SSv2 as the main benchmark in a perspective view of validating the importance of temporal modeling.

\vspace{0.03in}
\noindent{\bf Baselines.}
We consider recent Video Transformers as baselines: TimeSformer~\cite{bertasius2021times} with divided space and time attentions, Motionformer~\cite{patrick2021motions} with  trajectory attention, and X-ViT~\cite{bulat2021xvit} with space-time mixing attention.
All Video Transformers in our experiments are based on  ViT-B/16~\cite{dosovitskiy2021vit} (86M parameters), which consists of 12 transformer blocks with 768 embedding dimension.
We denote our method built upon an existing method by \mbox{“+ TIME”}, \emph{e.g.,} TimeSformer + TIME.
For Figure \ref{fig:fig1a} and \ref{fig:fig1b} and Table \ref{table1} in Section \ref{sec:example},
we use publicly available pre-trained models and validate their ability of temporal modeling.
{We remark that Figure \ref{fig:fig1a} shows the importance of temporal order information in the temporal classes, but such information vanishes as the model layers deeper as shown in Figure \ref{fig:fig1b}.}

\begin{table*}[t]
\centering
\begin{tabular}{lc|ccc}
\toprule
Dataset & Baseline & Baseline + ${\mathcal{L}}_I^{\text{order}}$ & Baseline + ${\mathcal{L}}_I^{\text{debias}}$ & Baseline + ${\mathcal{L}}_I^{\text{TIME}}$ \\
\midrule
Original {\color{blue}$\uparrow$}    & 77.3 & 82.0 ($+$4.7) & 79.0 ($+$1.7) & \textbf{83.3 ($+$6.0)}\\
Only-FG {\color{blue}$\uparrow$}     & 50.3 & 54.2 ($+$3.9) & 52.7 ($+$2.4) & \textbf{58.9 ($+$8.6)}\\
Mixed-Same {\color{blue}$\uparrow$}  & 68.6 & 72.5 ($+$3.9) & 69.7 ($+$1.1) & \textbf{74.0 ($+$5.4)}\\
Mixed-Rand {\color{blue}$\uparrow$}  & 43.7 & 48.4 ($+$4.7) & 45.1 ($+$1.4) & \textbf{51.0 ($+$7.3)}\\
Mixed-Next {\color{blue}$\uparrow$}  & 39.9 & 43.6 ($+$3.7) & 40.6 ($+$0.7) & \textbf{46.4 ($+$6.5)}\\
\midrule
BG-Gap {\color{red}$\downarrow$}    & 24.8 & 24.1 ($-$0.7) & 24.6 ($-$0.2) & \textbf{23.0 ($-$1.8)}\\
\bottomrule
\end{tabular}
\caption{Extension of our method to image classification models. Baseline denotes DeiT-Ti/16~\cite{touvron2020training}, and we train all models with 300 training epochs on ImageNet-9~\cite{xiao2021noise} and evaluate them for background shifts~\cite{xiao2021noise}. 
Original denotes original ImageNet-9 dataset, Only-FG, Mixed-Same, Mixed-Rand and Mixed-Next denote variation of ImageNet-9 by shifting image background; 
Only-FG remains only foregrounds and removing backgrounds, Mixed-Same, -Rand and -Next changes backgrounds to random backgrounds from the same, a random, and the next class, respectively.
BG-Gap denotes the difference between Mixed-Same and Mixed-Rand. Values in parenthesis are the performance difference between Baseline and Baseline incorporated with each loss term.}\label{table:imagenet}
\end{table*}

\vspace{0.03in}
\noindent{\bf Implementation details.} 
In our experiments, we unify different training details of the recent Video Transformers~\cite{bertasius2021times,patrick2021motions,bulat2021xvit}, and re-implement all the baselines on our setups for a fair comparison.\footnote{Code is available at \url{https://github.com/alinlab/temporal-selfsupervision}.}
Specifically, we fine-tune all the models from ImageNet~\cite{deng2009imagenet} pre-trained weights of ViT-B/16~\cite{dosovitskiy2021vit} for 35 training epochs with Adamw optimizer~\cite{loshchilov2018adamw} and learning rate of 0.0001 and a batch size of 64. For data augmentation, we follow RandAugment~\cite{cubuk2020randaugment} policy of \citet{patrick2021motions}. 
We use the spatial resolution of $224\times224$ with patch size of $16\times16$, and 
eight frame input videos under the same 1 × 16 × 16 tokenization method, including Motionformer.
We set all the loss weights to be 1 (\emph{i.e.,} $\lambda^{\text{order}}=\lambda^{\text{debias}}=\lambda^{\text{flow}}=1$) 
unless stated otherwise.

\subsection{Temporal modeling on SSv2}
In this section, we evaluate our method on the SSv2 benchmark using eight frame input videos by incorporating with the state-of-the-art Video Transformers; TimeSformer~\cite{bertasius2021times}, Motionformer~\cite{patrick2021motions}, and X-ViT~\cite{bulat2021xvit}. 
Under the same training details; \emph{e.g.,} optimizer, training scheduling, augmentation policy and tokenization, as shown in Table \ref{table:ssv2},
we found that our method, TIME, consistently improves all the backbone architectures with a large margin.
For example, TimeSformer + TIME and X-ViT + TIME achieve 1.6\% and 3.4\% higher top-1 accuracies compared to their baselines TimeSformer (62.1\%) and X-ViT (60.1\%), respectively.
These results not only show the effectiveness of TIME but also demonstrate the high architectural compatibility of TIME and allow us to conjecture that ours can overcome failure modes in the Video Transformers.
One can further verify the advantage of our scheme from the provided qualitative examples in Figure \ref{fig:fig3}. Here, with better temporal modeling from the proposed self-supervised tasks, the model can capture the temporal dynamics in a better way. More examples and details are in Appendix \ref{appendix:example}.

\subsection{Ablation study}\label{sec:ablation} 
In this section, we perform an ablation study to understand further how TIME works. Specifically, we perform TimeSformer + TIME using eight frame input videos varying loss components 
(in Eq. (\ref{eq:time})) to demonstrate their effectiveness; (a) learning temporal order of frames $\mathcal{L}^{\text{order}}$, (b) debiasing spatial dynamics $\mathcal{L}^{\text{debias}}$, and (c) learning temporal flow direction of tokens $\mathcal{L}^{\text{flow}}$.
To further validate the importance of temporal information in various experimental setups,
we train video models with the same training details on the full SSv2 dataset, the Temporal and Static subsets
(see Appendix~\ref{app:temporal_static} for their configurations). We report the test accuracies of both original and shuffled videos and their gap as a measurement of 
avoiding the spatial bias.

Table \ref{table:lossweight} summarizes the results: 
our loss components have an orthogonal contribution to the overall improvements in the model generalization (\emph{i.e.}, Original),
and TimeSformer + TIME (\emph{i.e.}, using all components $\mathcal{L}^{\text{order}}$, $\mathcal{L}^{\text{debias}}$, and $\mathcal{L}^{\text{flow}}$) consistently improves 
the performances of Original and Gap on all benchmarks.
For example, our method in the last row achieves the best score, which are 1.6\%, 5.3\%, and 2.8\% higher Original accuracies on the SSv2, the Temporal and Static subsets, respectively.
TimeSformer + TIME also largely surpasses TimeSformer in the metric of Gap by achieving 17.6, 40.2, and 17.6 improvements on the SSv2, the Temporal and Static subsets, respectively.

In addition, Table \ref{table:lossweight} shows that 
our loss components contribute more significantly when the dataset requires more temporal understanding
(\emph{e.g.}, the Temporal subset). 
For example, our method in the last row achieves more significant improvements in the metrics of Original and Gap on the Temporal subset compared to the scores on the SSv2 dataset and Static subset.
Interestingly, except for the last row, the performances of Shuffled on the Static subset are often close to the Original ones.
It arguably results from the static classes that allow video models to predict class labels without understanding temporal information,
while our scheme of debiasing spurious correlation (\emph{i.e.,} $\mathcal{L}^{\text{order}}$ and $\mathcal{L}^{\text{debias}}$) and enhancing temporal correlation (\emph{i.e.,} $\mathcal{L}^{\text{flow}}$)
can lead video models to alleviate spatial bias and learn effective temporal modeling.
We believe that an elaborate design for utilizing such property might further improve the video understanding, and we leave it for future work.

\subsection{Extension to image classification}\label{sec:image}
In this section, we demonstrate an extension of our self-supervised idea of debiasing to image classification models, \emph{e.g.,} Vision Transformers~\cite{dosovitskiy2021vit,touvron2020training}, to reduce the spurious correlation learned from the image backgrounds.
For adaptation, our goal is to replace (a) learning temporal order of frames with spatial order of patches and (b) debiasing spatial dynamics with image backgrounds.
For brevity, we again use $f_\theta$ to denote the encoder of Vision Transformers parameterized by $\theta$
where $f_\theta^\mathtt{[CLS]}(\mathbf{x})$ and $f_\theta^{(i)}(\mathbf{x})$ are the final representations of the $\mathtt{[CLS]}$ token and the $i$-th patch image, respectively. 
Then, (a) learning spatial order of patches can be written as follow:
\begin{align}
{\mathcal{L}}_{I}^{\text{order}}(\mathbf{x})&:= \frac{1}{s}\sum_{i=1}^s \text{CE}({g}^{\text{order}}_\theta(f_\theta^{(i)}(\mathbf{x})),\mathbf{y}^{(i)}), 
\end{align}
where $g^{\text{order}}_\theta$ is an linear classification head. On the other hand, (b) debiasing image backgounds objective also can be written as follow:
\begin{align}
{\mathcal{L}}_{I}^{\text{debias}}(\widetilde{\mathbf{x}}):= 
\text{KL}(\mathcal{U}(\mathbf{y})
    \big\| \mathtt{Softmax}(g_\theta(f_\theta^\mathtt{[CLS]}(\mathbf{\widetilde{x}})))),
\end{align}
where $g_\theta$ is an linear classification head, and $\tilde{\mathbf{x}}$ is a sequence of randomly shuffled patches to reduce the effect of backgrounds. Then, adapted loss objectives ${\mathcal{L}}_I^{\text{TIME}}$ for image classification models can be written as follows:
\begin{align}
{\mathcal{L}}_I^{\text{TIME}}(\mathbf{x},\mathbf{\tilde{x}})&:=\lambda_I^{\text{order}}{\mathcal{L}}_I^{\text{order}}(\mathbf{x})+\lambda_I^{\text{debias}}{\mathcal{L}}_I^{\text{debias}}(\tilde{\mathbf{x}}), \label{eq:image_time}
\end{align}
where $\lambda_I^{\text{order}}$ and $\lambda_I^{\text{debias}}$ are hyperparameters. In Table \ref{table:imagenet}, we also simply use $\lambda_I^{\text{order}}=\lambda_I^{\text{debias}}=1$.

Somewhat surprisingly, we found that adapted TIME loss (\ref{eq:image_time}) enhances the model generalization and robustness to background shifts when evaluated on Backgrounds Challenge\footnote{\scriptsize \url{https://github.com/MadryLab/backgrounds_challenge}} as shown in Table \ref{table:imagenet}.
Specifically, we train DeiT-Ti/16~\cite{touvron2020training} with 300 training epochs on ImageNet-9 \cite{xiao2021noise} dataset, which contains 9 super-classes (370 classes in total) of ImageNet~\cite{deng2009imagenet} for both background shifts experiments~\cite{xiao2021noise}.
For example, Only-FG replaces backgrounds with the black, and Mixed-Same, Mixed-Rand, and Mixed-Next replace backgrounds with random backgrounds from the same, a random, and the next class, respectively, for disentangling foregrounds and backgrounds of the images.

Table \ref{table:imagenet} summarizes the results on the background shifts: 
our method consistently and significantly improves Baseline on overall benchmarks;
\emph{e.g.,} Baseline + ${\mathcal{L}}_I^{\text{TIME}}$ not only improves the top-1 accuracy of Baseline
from 77.26\% to 83.26\% on the original dataset but also from 50.27\% to 58.91\% on the Only-FG benchmark in the background shifts.
Also, Table \ref{table:imagenet} shows that ${\mathcal{L}}_I^{\text{order}}$ and ${\mathcal{L}}_I^{\text{debias}}$ have an orthogonal contribution to the overall improvements
for alleviating background bias. For example, ${\mathcal{L}}_I^{\text{order}}$ improves Baseline from 50.27\% to 54.15\%, and ${\mathcal{L}}_I^{\text{debias}}$ improves it again from 54.15\% to 58.91\% on the Only-FG benchmark.
Furthermore, comparing performance gains from each loss component and the combined one, it is worth noting that we confirm the remarkable synergy of spatial order prediction and background debiasing for robust image recognition in most benchmarks. This superiority of our method on background shifts shows its merits come from a widely applicable self-supervised idea for the vision domain.
\section{Conclusion}\label{conclusion}
We propose simple yet effective temporal self-supervision tasks (TIME) for improving video representations by capturing temporal dynamics of video data.
Our key observation is that the existing Video Transformers do ineffective temporal modeling; \emph{e.g.,} learning spurious correlation on spatial dynamics and vanishing temporal order information as the model layers deeper. 
In order to learn effective temporal modeling, we 
train the model to output low-confident predictions for temporally violated video data (\emph{e.g.,} randomly shuffled video), 
and to predict both the correct temporal order of video frames and the temporal flow direction of video tokens.
Through the extensive experiments, 
we highlight the importance of debiasing the spurious correlation of visual transformers with respect to the temporal or spatial dynamics.
We believe that our work would provide insights toward the under-explored yet important problem.

\noindent{\bf Acknowledgements.}
We thank Seong Hyeon Park for providing helpful feedback and suggestions.
This work was mainly supported by Institute of Information \& communications Technology Planning \& Evaluation
(IITP) grant funded by the Korea government (MSIT) (No.2021-0-02068, Artificial Intelligence
Innovation Hub; No.2019-0-00075, Artificial Intelligence Graduate School Program (KAIST)). This
work was partly experimented on the NAVER Smart Machine Learning (NSML) platform (Sung
et al., 2017; Kim et al., 2018). This work was partly supported by KAIST-NAVER Hypercreative AI
Center.


\bibliography{icml2022}
\bibliographystyle{icml2022}


\newpage
\appendix
\onecolumn
\section{Configurations of Temporal and Static subsets in Something-Something-v2} \label{app:temporal_static}
By following \citet{sevilla2021only}, we use the categorization of 18 temporal classes and 16 static classes of the Something-Sonething-v2 (SSv2) dataset~\cite{goyal2017something}.

\vspace{0.01in}
\noindent{\bf Temporal classes.} 
``Turning something upside down", 
``Approaching something with your camera", 
``Moving something away from the camera", 
``Moving away from something with your camera", 
``Moving something towards the camera", 
``Lifting something with something on it",
``Moving something away from something", 
``Moving something closer to something", 
``Uncovering something", 
``Pretending to turn something upside down", 
``Covering something with something", 
``Lifting up one end of something, then letting it drop down", 
``Lifting something up completely without letting it drop down", 
``Moving something and something closer to each other", 
``Moving something and something away from each other", 
``Lifting something up completely, then letting it drop down", 
``Stuffing something into something", 
``Moving something and something so they collide with each other".

Specifically, we use 6 temporal classes of
``Lifting something up completely without letting it drop down", 
``Lifting something up completely, then letting it drop down", 
``Lifting up one end of something, then letting it drop down", 
``Moving something and something closer to each other", 
``Moving something and something away from each other", 
and ``Moving something and something so they collide with each other" classes as the Temporal subset.

\vspace{0.01in}
\noindent{\bf Static classes.} 
``Folding something", 
``Turning the camera upwards while filming something", 
``Holding something next to something, Pouring something into something", 
``Pretending to throw something", 
``Squeezing something", 
``Holding something in front of something", 
``Touching (without moving) part of something",
``Lifting up one end of something without letting it drop down", 
``Showing something next to something",
``Poking something so that it falls over", 
``Wiping something off of something", 
``Scooping something up  with something", 
``Letting something roll down a slanted surface", 
``Sprinkling something onto something",
``Pushing something so it spins", 
``Twisting (wringing) something wet until water comes out".
We use all the above 16 static classes as the Static subset.

\vspace{-0.03in}
\section{Visualization of learned video models via GradCAM}\label{appendix:example}
\begin{figure}[b]
\begin{center}
    {
    \subfigure[Examples from class \emph{moving something and something closer to each other}.]
        {
        \includegraphics[width=0.95\textwidth]{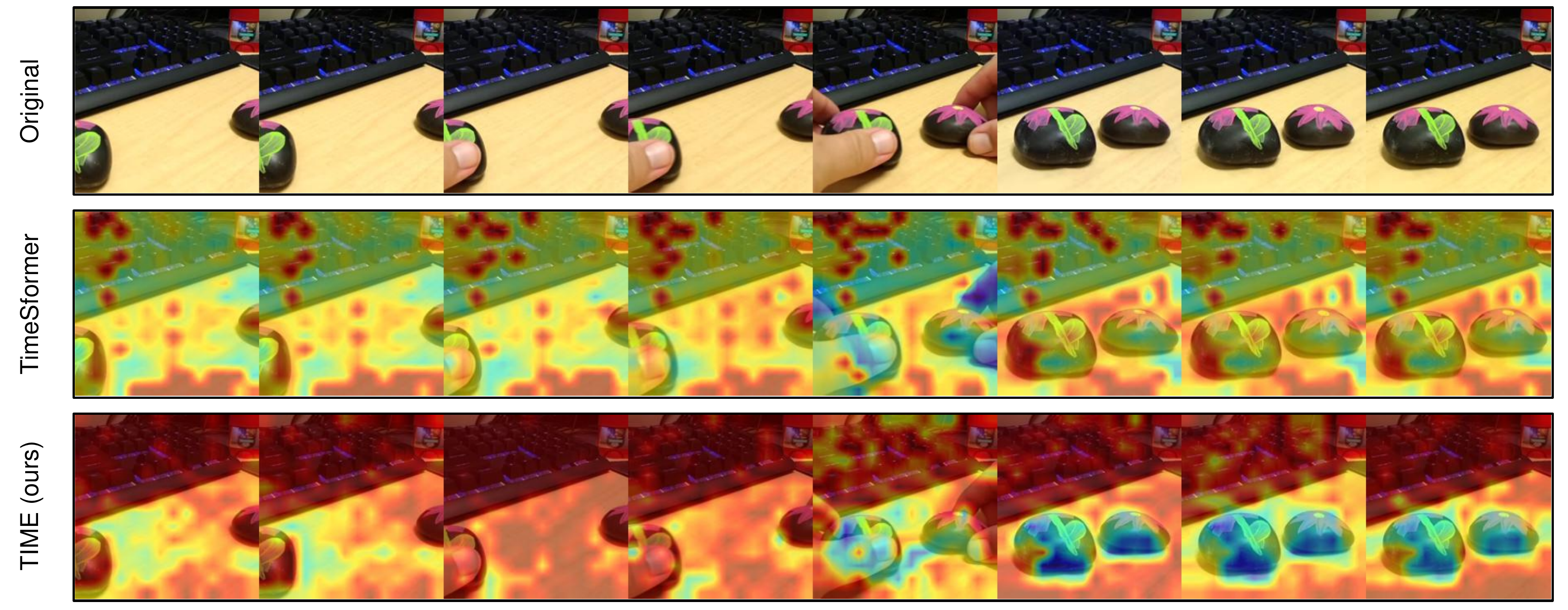}
        \label{appendix_fig1a}
        }
    \subfigure[Examples from class \emph{moving something and something so they collide with each other}.]
        {
        \includegraphics[width=0.95\textwidth]{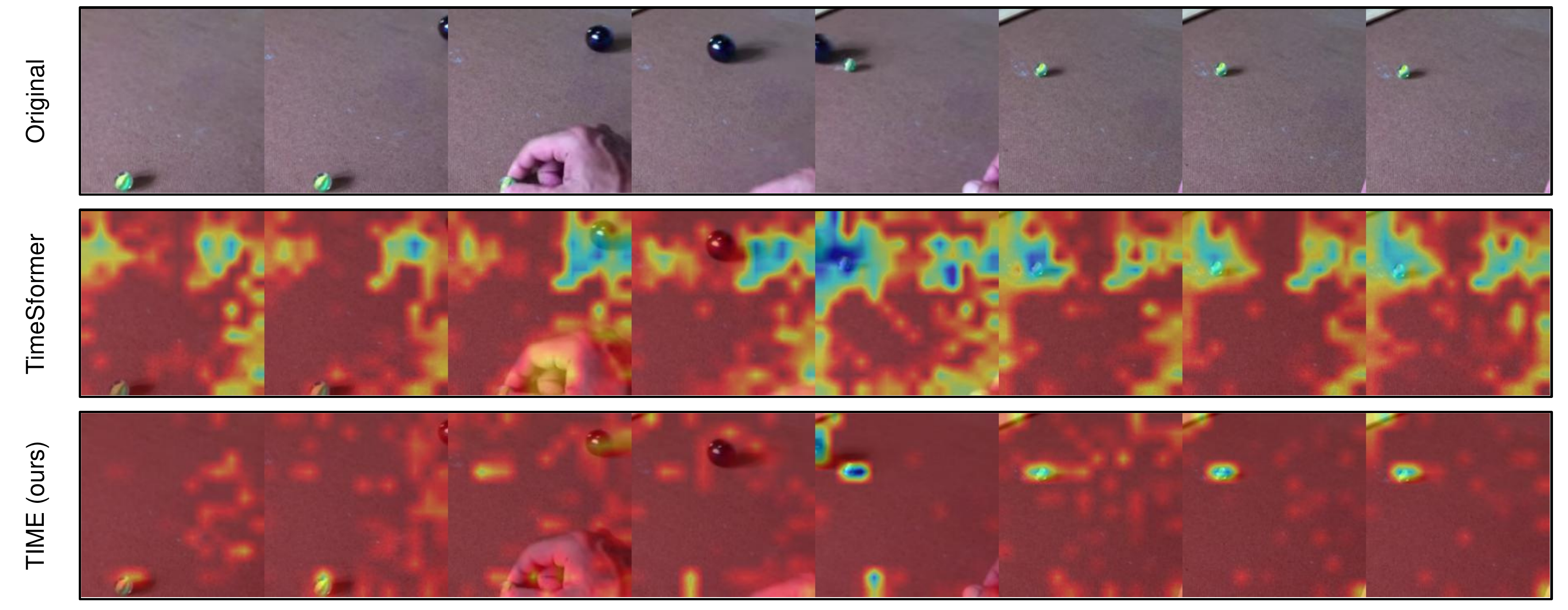}
        \label{appendix_fig1b}
        }
    \subfigure[Examples from class \emph{turning something upside down}.]
        {
        \includegraphics[width=0.95\textwidth]{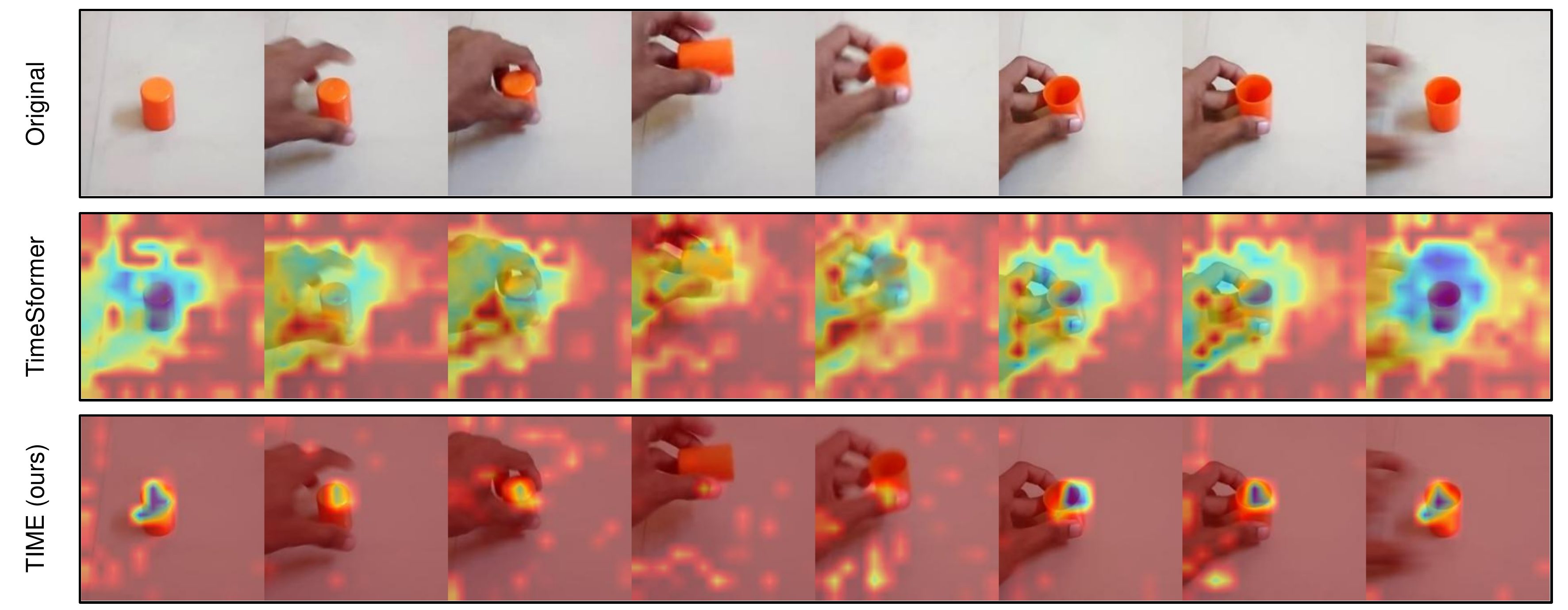}
        \label{appendix_fig1c}
        }
    }
\end{center}
\caption{More qualitative examples on SSv2 test dataset using GradCAM.}
\label{fig:appendix_examples}
\end{figure}

In this section, we present details of qualitative results in Figure \ref{fig:fig3}, and then provide more examples in Figure \ref{fig:appendix_examples}.
To apply GradCAM \cite{selvaraju2017grad}, which was originally developed based on CNNs, 
we use its adaptation for Vision Transformers provided by the original authors.\footnote{\scriptsize \url{https://github.com/jacobgil/pytorch-grad-cam}} 
As the code is originally proposed for image data, we slightly modify it for video data by considering one more dimension for the temporal dimension (\emph{i.e.}, frames).
Here, one can again identify that our method successfully improves the existing Video Transformers with the proposed temporal self-supervised tasks; \emph{e.g.}, focusing on the movements of objects (see Figure \ref{appendix_fig1a} and \ref{appendix_fig1b}) or the turning object (see Figure \ref{appendix_fig1c}).

\vspace{-0.03in}
\section{Temporal modeling of TIME in training from scratch}\label{app:scratch}
As many recent Video Transformers depend on pre-trained weights from a large-scale image dataset (\emph{e.g.}, ImageNet~\citep{deng2009imagenet}) for better performances, the pre-trained weights could be one of the reasons for spatial bias in the video models. 
However, even without the pre-trained weights, we empirically found that our method, TIME, still significantly improves TimeSformer from 39.4\% to 64.4\% in training from scratch on the Temporal subset using eight frame input videos. 
Again, we remark that static classes in the video datasets can also make video models biased toward learning spatial dynamics. Still, our method can lead them to alleviate spatial bias and learn temporal modeling better.

\vspace{-0.03in}
\section{Alternative for keeping temporal information}\label{app:arch_modify}
Here, we explore an alternative for keeping temporal information in Video Transformers. 
Specifically, 
we modify TimeSformer~\cite{bertasius2021times} to repeatedly re-add their temporal positional encodings before each block to maintain the temporal order of video frames and then train them on the Temporal subset.
Interestingly, we observed that the re-adding one still loses temporal information at the
final layer (27.8\%), as shown in Figure~\ref{fig:fig1b}. 
Furthermore, we found that learning temporal order $\mathcal{L}^{\text{order}}$ (\ref{eq_tempo}) could improve the re-adding one from 88.2\% to 90.0\% on the Temporal subset using 16 frame input videos.\footnote{We note the baseline (\emph{i.e.}, the original TimeSformer) and the re-adding one + TIME achieve 87.2\% and 90.5\%, respectively.} 
These results show that architectural modifications for solely guiding temporal
information may be limited; however, investigating architectural advances to keep temporal information would be a meaningful future direction.


\end{document}